\documentclass{article}
\usepackage{arxiv}

\usepackage[T1]{fontenc}
\usepackage[normalem]{ulem}
\usepackage{latexsym}
\usepackage{amsmath, amssymb}
\usepackage{mathtools}
\mathtoolsset{showonlyrefs}
\usepackage{siunitx}
\usepackage{algorithm}
\usepackage{algpseudocode}
\algrenewcommand\algorithmicrequire{\textbf{Input:}}
\algrenewcommand\algorithmicensure{\textbf{Output:}}
\usepackage{wrapfig}
\usepackage{graphicx}
\usepackage{xcolor}
\usepackage{glossaries}
\usepackage{enumitem}
\usepackage{booktabs}
\usepackage{adjustbox}
\usepackage{tabularx}
\usepackage{multirow}
\usepackage{caption} 
\usepackage{subcaption} 
\usepackage{tikz}
\usepackage{comment}
\usepackage{xurl}
\usepackage{hyperref}
\hypersetup{%
  breaklinks=true,%
  linktocpage=true,
  colorlinks=true,%
  linkcolor=black,%
  citecolor=purple,%
  filecolor=blue,%
  urlcolor=blue,%
  pdftitle={},%
  pdfsubject={},%
  pdfauthor={},%
  pdfkeywords={},%
  pdfproducer={}%
  }%
  
\usepackage[textwidth=2.5cm, textsize=scriptsize]{todonotes}

\usepackage{pgfplots}
\usepgfplotslibrary{groupplots}
\pgfplotsset{compat=1.18}
\usetikzlibrary{calc}

\newcommand{\var}{\mathrm{var}}
\newcommand{\E}{\mathbb{E}}
\DeclareMathOperator{\train}{\text{\texttt{train}}}
\DeclareMathOperator{\unlearn}{\text{\texttt{unlearn}}}
\DeclareMathOperator{\evaluate}{\text{\texttt{evaluate}}}
\newcommand{\seed}{\text{\texttt{seed}}}

\newcommand{\trainSeed}{s}
\newcommand{\TrainSeed}{S}
\newcommand{\unlearnSeed}{\tilde{s}}
\newcommand{\UnlearnSeed}{\tilde{S}}
\newcommand{\nTrainSeeds}{I}
\newcommand{\nUnlearnSeeds}{J}

\newcommand{\trainSubscript}{\mathrm{train}}
\newcommand{\unlearnSubscript}{\mathrm{unlearn}}
\newcommand{\overallSubscript}{\mathrm{overall}}

\newcommand{\varTrain}{\smash{\sigma_{\trainSubscript}^2}}
\newcommand{\varUnlearn}{\smash{\sigma_{\unlearnSubscript}^2}}
\newcommand{\varOverall}{\smash{\sigma_{\overallSubscript}^2}}

\newcommand{\varTrainEst}{\smash{\hat{\sigma}_{\trainSubscript}^2}}
\newcommand{\varUnlearnEst}{\smash{\hat{\sigma}_{\unlearnSubscript}^2}}

\newcommand{\sdTrainEst}{\smash{\hat{\sigma}_{\trainSubscript}}}
\newcommand{\sdUnlearnEst}{\smash{\hat{\sigma}_{\unlearnSubscript}}}
\newcommand{\sdOverallEst}{\smash{\hat{\sigma}_{\overallSubscript}}}

\newcommand{\icc}{\mathrm{ICC}}
\newcommand{\iccEst}{\smash{\widehat{\icc}}}

\newcommand{\costTrain}{\smash{c_{\trainSubscript}}}
\newcommand{\costUnlearn}{\smash{c_{\unlearnSubscript}}}

\newcommand{\metric}{Z}
\newcommand{\metricAvg}{\mathcal{\metric}}

\newcommand{\ciWidth}{W}

\newcommand{\fisher}{\mathcal{F}}

\newacronym{MIA}{MIA}{membership inference attack}
\newacronym{LLM}{LLM}{large language model}
\newacronym{SSD}{SSD}{selective synaptic dampening}
\newacronym{LFSSD}{LFSSD}{loss-free selective synaptic dampening}
\newacronym{FOLTR}{FOLTR}{federated online
learning-to-rank}
\newacronym{ICC}{ICC}{intraclass correlation coefficient}
\newacronym{CI}{CI}{confidence interval}

\usepackage{graphicx}
\usepackage{color}

\begin{document}
\title{On the importance of multiple training seeds for evaluating machine unlearning} 

\author{
 Jamie Lanyon \\
  Department of Computer Science\\
  School of Science\\
  Loughborough University\\
  \texttt{j.lanyon@lboro.ac.uk} \\
   \And
 Axel Finke \\
    School of Mathematics\\
    Statistics and Physics\\
    Newcastle University\\
    \texttt{axel.finke@newcastle.ac.uk} \\
  \And
 Petros Andreou \\
  Department of Computer Science\\
  School of Science\\
  Loughborough University\\
  \texttt{p.andreou2@lboro.ac.uk} \\
  \And
 Georgina Cosma \\
  Department of Computer Science\\
  School of Science\\
  Loughborough University\\
  \texttt{g.cosma@lboro.ac.uk} \\
}      

\glsunset{SSD}
\glsunset{LFSSD}

\maketitle

\begin{abstract}
Machine unlearning aims to remove the influence of certain data points from a trained model without costly retraining. Most practical unlearning algorithms are only approximate and their performance can only be assessed empirically. Common practice is to run unlearning algorithms multiple times independently (i.e., using multiple \emph{unlearning seeds}) starting from the same trained model (i.e., using only a single \emph{training seed}). In image-classification experiments, this practice can give non-representative results as unlearning performance can be sensitive to the choice of training seed. This is particularly relevant for deterministic unlearning methods which always produce the same result when started from the same trained model.
Further experiments on federated learning-to-rank, and large language models confirm that this issue extends beyond image classification. We also explain why increasing the number of unlearning seeds cannot generally compensate for the lack of multiple training seeds. Finally, we give guidance on how to select the number of training and unlearning seeds.

\end{abstract}

\section{Introduction}

The targeted removal of the influence of certain training data on the parameters of machine-learning models \cite{Machine_Unlearning_Cao} is often needed, e.g., for regulatory compliance (e.g., GDPR), elimination of biased or corrupted samples, or mitigation of data poisoning attacks. These concerns extend to information retrieval systems \cite{hou2024neural}. However, the exact (`gold-standard') solution to the machine-unlearning  problem, which involves retraining the model from scratch without the to-be-forgotten data (called \emph{Retrain}), is typically infeasible. Instead, a number of approximate unlearning methods have been developed to enable models to forget specific data whilst maintaining overall performance, e.g., \emph{Random Labels} \cite{GolatkarEternalsunshineoftheSpotless202}, \emph{UNSIR} \cite{Tarun_2024_UNSIR}, \emph{Bad Teacher} \cite{chundawat2023badteachinginduceforgetting}, \emph{\gls{SSD}} \cite{Foster_Schoepf_Brintrup_2024} or \emph{\gls{LFSSD}} \cite{foster2024loss}. 

Unlearning algorithms are often difficult to analyse mathematically. Their evaluation must then be done empirically. This calls for empirical experiments to be designed carefully so as to provide confidence that their findings generalise beyond the specific setting of the study (e.g., architecture or dataset). Common practice \cite{chundawat2023badteachinginduceforgetting,Chundawat_2023,GolatkarEternalsunshineoftheSpotless202,Tarun_2024_UNSIR,Foster_Schoepf_Brintrup_2024,foster2024loss} is to run \emph{stochastic} unlearning algorithms (e.g., UNSIR, Random Labels or Bad Teacher) multiple times independently -- each using a different (pseudo) \emph{random-number generator} seed, henceforth called an \emph{unlearning seed}, to account for their inherent variability. One then assesses the empirical distribution (across unlearning seeds) of unlearning metrics such as accuracy on the data to be forgotten (\emph{forget accuracy}) or accuracy on the remaining data (\emph{retain accuracy}). In contrast, \emph{deterministic} unlearning algorithms, such as \gls{SSD} and LFSSD, are only run once \cite{Foster_Schoepf_Brintrup_2024,foster2024loss} because they always return the same unlearned model (and hence the same values of any unlearning metric) when started from the same trained model -- irrespective of the unlearning seed.

While comparisons across multiple unlearning seeds are common, many works use a single \emph{training seed,} i.e., a single trained model as the starting point of unlearning. In this work, we demonstrate that this evaluation strategy can lead to unreliable unlearning evaluations (see Figure~\ref{fig:intro_figure_trimmed}). Our contributions are:
\begin{enumerate}
    \item In Section~\ref{sec:experiments}, we numerically demonstrate (in the context of image classification) that the common single-training-seed based evaluation of unlearning methods can produce non-representative results, even for a fixed architecture and dataset. This issue extends beyond image classification to unlearning two other tasks: \emph{\gls{FOLTR}} and \emph{\glspl{LLM}} as we show in Appendices~\ref{app:foltr-details} and \ref{app:LLM-details}.
    \item In Section~\ref{subsec:variance_decomposition_analysis}, we explain why using multiple unlearning seeds typically cannot compensate for the lack of multiple training seeds.
    \item In Section~\ref{sec:compute-allocation}, we provide guidance on allocating computational resources between training-seed and unlearning-seed repetitions.
\end{enumerate}
\begin{figure}[!htb]
    \centering
    \input{tikz/intro_figure_trimmed.tex}
    \vspace{-2ex}
    \caption{Variability unlearning performance in the image classification scenario from Section~\ref{sec:experiments}. 
    Each boxplot visualises $15$ estimates of the standard deviation of the retain and forget accuracies (in percentage points): overall ($\sdOverallEst$), across training seeds ($\sdTrainEst$), and across unlearning seeds ($\sdUnlearnEst$). These $15$ estimates consist of three unlearning scenarios: CIFAR-100 (full-class), CIFAR-20 (full-class) and CIFAR-20 (sub-class); each with five forget classes. Each of these $15$ estimates is based on 75 training seeds; each training seed is combined with ten unlearning seeds. This figure illustrates that unlearning performance can be sensitive to the training seed so that accurate unlearning evaluation requires averaging across multiple training seeds. In contrast, variability across unlearning seeds (and of the Retrain baseline) is typically of a smaller magnitude. Second axis truncated for clarity -- see Appendix~\ref{app:subsec:extended_version_of_intro_figure} for a full version of this figure.}
    \label{fig:intro_figure_trimmed}
\end{figure}
Our work extends findings in the literature. 

Cadet et al.~\cite{cadet2025deep} found variability of unlearning performance across ten training seeds even with the same hyper-parameters and data splits. Our study (a) is larger, using 75 training seeds in the image classification scenarios; (b) explains why multiple unlearning seeds cannot compensate for a lack of multiple training seeds; (c) includes guidance on the number of training/unlearning seeds to use. 

Triantafillou et al.~\cite{triantafillou2024makingprogressunlearningfindings} noted that a valid statistical analysis would require averaging unlearning metrics across training seeds. However, their experiments use single training seeds and focus primarily on the variability across unlearning seeds and of the Retrain baseline. As shown in Figure~\ref{fig:intro_figure_trimmed}, we observe that these variabilities are typically smaller than the variability attributable to training seeds. Finally, our additional experiments on \gls{FOLTR} and \glspl{LLM} go beyond the image classification scenario considered in \cite{cadet2025deep,triantafillou2024makingprogressunlearningfindings}.

In this work, we corrected an implementation issue in the reference \gls{SSD} and \gls{LFSSD} code which biases their parameter importance scores and causes non-deterministic behaviour across unlearning seeds. To distinguish the two implementations, we will refer to the original implementation used in \cite{Foster_Schoepf_Brintrup_2024,foster2024loss} as ``SSD/LFSSD (orig.)'' and to our updated implementation as ``SSD/LFSSD (det.)''. We note that both implementations share the same hyperparameters. Full details are provided in Appendix~\ref{sec:det_ssd}. 
    
\section{Methods}

\subsection{Generic Evaluation Procedure}

This section describes the evaluation procedure used to compare unlearning methods across multiple training and unlearning seeds, and presents a variance-decomposition analysis that motivates the use of multiple training seeds. Algorithm~\ref{alg:generic_evaluation_of_unlearning} outlines a generic empirical evaluation of some unlearning method for a given architecture and dataset. 
\begin{algorithm}
\caption{Generic empirical evaluation of unlearning}
\label{alg:generic_evaluation_of_unlearning}
\begin{algorithmic}[1]
\Require Number of distinct training seeds: $\nTrainSeeds \geq 1$.
\Require Number of distinct unlearning seeds per training seed: $\nUnlearnSeeds \geq 1$.
\Require Distinct training seeds $\trainSeed_1, \dotsc, \trainSeed_{\nTrainSeeds}$.
\Require Distinct unlearning seeds $\unlearnSeed_{i,1}, \dotsc, \unlearnSeed_{i,\nUnlearnSeeds}$, for each $1 \leq i \leq \nTrainSeeds$.
\For {$i = 1, \dotsc, \nTrainSeeds$}
  \State \label{alg:generic_evaluation_of_unlearning:train} $M_i \leftarrow \train(\seed = \trainSeed_i)$ \Comment{trained model}
  \For {$j = 1, \dotsc, \nUnlearnSeeds$}
    \State \label{alg:generic_evaluation_of_unlearning:unlearn} $U_{i,j} \leftarrow \unlearn(M_i, \seed = \unlearnSeed_{i,j})$ \Comment{unlearned model}
    \State \label{alg:generic_evaluation_of_unlearning:evaluate} $\metric_{i,j} \leftarrow \evaluate(U_{i,j})$ \Comment{value of unlearning metric}
  \EndFor
\EndFor
\Ensure $\metricAvg \leftarrow (\nTrainSeeds \nUnlearnSeeds)^{-1} \sum_{i=1}^{\nTrainSeeds} \sum_{j=1}^{\nUnlearnSeeds} \metric_{i,j}$ \Comment{average value of unlearning metric}
\end{algorithmic}
\end{algorithm}
Here, the model is trained $\nTrainSeeds$ times independently using a different training seed; each of the trained models is then a starting point for $\nUnlearnSeeds$ (conditionally) independent runs of the unlearning method. Finally, the unlearning performance is evaluated by calculating the average value $\metricAvg$  of some unlearning metric (e.g., forget accuracy or retain accuracy) across all $\nTrainSeeds \nUnlearnSeeds$ unlearned models. Note that to ensure independence across training seeds, it is important that all $\nTrainSeeds$ training seeds are mutually distinct and that all $\nTrainSeeds\nUnlearnSeeds$ unlearning seeds are likewise mutually distinct. This can be achieved, e.g., by taking $\trainSeed_i \coloneqq i$ and $\unlearnSeed_{i,j} \coloneqq \nUnlearnSeeds (i - 1) + j$, for all $i \in \{1, \dotsc, \nTrainSeeds\}$ and $j \in \{1, \dotsc, \nUnlearnSeeds\}$.

A \emph{widespread practice} is to evaluate unlearning using Algorithm~\ref{alg:generic_evaluation_of_unlearning} with
 \begin{itemize}
     \item only $\nTrainSeeds = 1$ training seed;
     \item $\nUnlearnSeeds \geq 1$ unlearning seeds (e.g., $\nUnlearnSeeds = 3$ \cite{GolatkarEternalsunshineoftheSpotless202}, $\nUnlearnSeeds = 5$ \cite{chundawat2023badteachinginduceforgetting}, $\nUnlearnSeeds = 10$ \cite{foster2024loss,Foster_Schoepf_Brintrup_2024}).
 \end{itemize}
The remainder of this work discusses why $\nTrainSeeds \gg 1$ is generally needed (``$\gg$'' should be read as ``much larger than'').

\subsection{Variance-Decomposition Analysis} 
\label{subsec:variance_decomposition_analysis}

Let $\metric = h(\TrainSeed, \UnlearnSeed)$ be the random value of the chosen unlearning metric (e.g., forget or retain accuracy), where $\TrainSeed$ and $\UnlearnSeed$ are the random and independent training and unlearning seeds, respectively, and where we assume that $h(\trainSeed, \unlearnSeed) \coloneqq \evaluate(\unlearn(\train(\seed = \trainSeed), \seed = \unlearnSeed))$ is deterministic (so that metric values are deterministic conditional on the unlearned model).

The overall variance of the unlearning metric can then be decomposed as
\begin{align}
    \varOverall \coloneqq \var[\metric] = \varTrain + \varUnlearn,
\end{align}
where
\begin{align}
    \varTrain & \coloneqq \var[\E(h(\TrainSeed, \UnlearnSeed) | \TrainSeed) ],\\
    \varUnlearn &\coloneqq \E[\var(h(\TrainSeed, \UnlearnSeed) | \TrainSeed)],
\end{align}
are the portions of $\var[\metric]$ attributable to training and unlearning, respectively. Hereafter, we assume that $\varOverall$ (and hence also $\varTrain$ and $\varUnlearn$) is finite. A standard variance-decomposition argument (Appendix~\ref{app:sec:variance_decomposition}) then gives
\begin{align}
  \var[\metricAvg]
  & = \frac{1}{\nTrainSeeds} \biggl(\varTrain + \frac{1}{\nUnlearnSeeds} \varUnlearn \biggr) = \frac{\varOverall}{\nTrainSeeds} \biggl(\icc + \frac{1}{\nUnlearnSeeds} (1 - \icc)\biggr), \label{eq:variance_decomposition}
\end{align}
where
\begin{align}
  \icc \coloneqq \frac{\varTrain}{\varTrain + \varUnlearn} = \frac{\varTrain}{\varOverall},
\end{align}
can be interpreted as the \emph{\gls{ICC}} if we treat the unlearning metric values for a given training seed but different unlearning seeds as observations belonging to the same group. 

If $\varOverall > 0$, then Equation~\refeq{eq:variance_decomposition} shows that consistent estimation of unlearning performance requires increasing the number of training seeds, $\nTrainSeeds$, whenever $\varTrain > 0$, i.e., $\icc > 0$. The above-mentioned widespread practice (i.e., increasing the number of unlearning seeds, $\nUnlearnSeeds$, without increasing $\nTrainSeeds$) could only consistently estimate unlearning performance if $\varTrain = 0$, i.e., $\icc = 0$.

Unfortunately, as our numerical results in Section~\ref{sec:experiments} show (see also Figure~\ref{fig:intro_figure_trimmed}), many unlearning methods have $\varTrain \gg 0$ (and hence $\icc \gg 0$) so that increasing the number of training seeds is needed for controlling $\var[\metricAvg]$.

\subsection{Choice of the Number of Training and Unlearning Seeds}
\label{sec:compute-allocation}

Let $\costTrain \in (0, \infty)$ (in seconds, say) be the computational cost of training (Line~\ref{alg:generic_evaluation_of_unlearning:train} of Algorithm~\ref{alg:generic_evaluation_of_unlearning}) and let $\costUnlearn \in (0, \infty)$ be the computational cost of unlearning and metric evaluation (Lines~\ref{alg:generic_evaluation_of_unlearning:unlearn}--\ref{alg:generic_evaluation_of_unlearning:evaluate} of Algorithm~\ref{alg:generic_evaluation_of_unlearning}). 

\paragraph{Fixed budget.}
Assume that we have available a fixed computational budget: $C \in [\costTrain + \costUnlearn, \infty)$. Equation~\ref{eq:variance_decomposition} then shows that we can reduce $\var[\metricAvg]$ by setting $\nUnlearnSeeds = 1$ and taking $\nTrainSeeds$ as large as the budget constraint allows. However, the cost of training is typically much larger than the cost of unlearning: $\costTrain \gg \costUnlearn$. Consequently, it can be beneficial to devote part of the computational budget to multiple unlearning runs ($\nUnlearnSeeds > 1$) per training seed as we now show.

By standard calculations, an optimal allocation $(\nTrainSeeds^*, \nUnlearnSeeds^*)$ of $(\nTrainSeeds, \nUnlearnSeeds)$ that minimises $\var[\metricAvg]$ under the budget constraint $C = \nTrainSeeds(\costTrain + \nUnlearnSeeds \costUnlearn)$ is given by
\begin{align}
  (\nTrainSeeds^*, \nUnlearnSeeds^*) 
  =
   \begin{cases}
    (\nTrainSeeds'(\nUnlearnSeeds'), \nUnlearnSeeds'), & \text{if $\varTrain > 0$ and  $\varUnlearn > 0$,}\\ 
    (\nTrainSeeds'(1), 1), & \text{if $\varTrain > 0$ and $\varUnlearn = 0$,}\\ 
    (1, \lfloor (C - \costTrain) / \costUnlearn \rfloor), & \text{if $\varTrain = 0$ and $\varUnlearn > 0$,}\\ 
    (1, 1), & \text{if $\varTrain = 0$ and $\varUnlearn = 0$,}
   \end{cases}
\end{align}
where
\begin{align}
    \nTrainSeeds'(\nUnlearnSeeds) \coloneqq \biggl\lfloor \frac{C}{\costTrain + \nUnlearnSeeds \costUnlearn}\biggr\rfloor, \qquad \text{and} \qquad
    \nUnlearnSeeds' \coloneqq \biggl\lceil \sqrt{
    \frac{1 -\icc}{\icc} \frac{\costTrain}{\costUnlearn} 
    }\biggr\rceil.
\end{align}

\paragraph{Fixed accuracy.} Assume that we wish to ensure that unlearning performance can be accurately estimated in the sense that a simple $(1-\alpha)$-\emph{\gls{CI}} for the true value of the unlearning metric achieves a width of at most $\ciWidth > 0$.

By standard sample-size calculations, an optimal allocation $(\nTrainSeeds^*, \nUnlearnSeeds^*)$ of $(\nTrainSeeds, \nUnlearnSeeds)$ that achieves (at most) the desired $(1-\alpha)$-\gls{CI} width $\ciWidth$ is then given by
\begin{align}
  (\nTrainSeeds^*, \nUnlearnSeeds^*) 
  =
   \begin{cases}
    (\nTrainSeeds''(\nUnlearnSeeds'), \nUnlearnSeeds'), & \text{if $\varTrain > 0$ and  $\varUnlearn > 0$,}\\ 
    (\nTrainSeeds''(1), 1), & \text{if $\varTrain > 0$ and $\varUnlearn = 0$,}\\ 
    (1, \lceil (2 z_{1-\frac{\alpha}{2}} / \ciWidth)^2 \varUnlearn \rceil), & \text{if $\varTrain = 0$ and $\varUnlearn > 0$,}\\ 
    (1, 1), & \text{if $\varTrain = 0$ and $\varUnlearn = 0$,}
   \end{cases}\label{eq:ci_sample_size}
\end{align}
where
\begin{align}
  \nTrainSeeds''(\nUnlearnSeeds) 
  & \coloneqq
  \biggl\lceil\biggl(\frac{2 z_{1-\frac{\alpha}{2}}}{\ciWidth}\biggr)^2 \varOverall \biggl(\icc + \frac{1}{\nUnlearnSeeds
  } (1 - \icc)\biggr)\biggr\rceil.
\end{align}
Note that the above \gls{CI} is based on a normal approximation which may be unreliable for proportion-valued metrics near 0 or 1 (e.g. forget accuracy); in this case, it might be prudent to consider remedies such as variance-stabilising (e.g. arcsine) transformations or Wilson intervals.

Of course, in practice, $\varTrain$ and $\varUnlearn$ (and hence $\icc$) are typically unknown and need to be replaced by suitable estimates $\varTrainEst$, $\varUnlearnEst$ (and hence $\iccEst$). In the next section, we provide such estimates for a number of unlearning methods in image classification scenarios; and while these estimates are specific to the tested scenarios, they caution against the widespread practice of evaluating unlearning using just $\nTrainSeeds = 1$ training seed.

\section{Experiments}
\label{sec:experiments}

In this section, we estimate $\varTrain$ and $\varUnlearn$ (and thus $\icc$) in computer vision scenarios. Additional tasks and dataset analyses are given in Appendix~\ref{app:foltr-details} (\gls{FOLTR}) and Appendix~\ref{app:LLM-details} (\glspl{LLM}). Appendix~\ref{app:code} provides links to full code for reproducing the results.

 As datasets, we chose CIFAR-100 and CIFAR-20 for full-class unlearning; and CIFAR-20 for sub-class unlearning \cite{chundawat2023badteachinginduceforgetting}. For the latter, the 100 classes in CIFAR-100 were grouped into 20 superclasses. We used a ResNet-18 architecture, randomly initialised and trained from scratch on each dataset. 

The full experimental setup is described in Appendix \ref{app:experimental-details-image}. 

For full-class unlearning, as shown in Table~\ref{tab:variance-components-full-class}, most unlearning methods exhibit higher training-seed variation than unlearning-seed variation. Consequently, Table~\ref{tab:optimal-seeds-full-class} indicates increasing the number of training seeds (rather than unlearning seeds) to reduce \gls{CI} widths. Results for sub-class unlearning, postponed to Appendix~\ref{app:experimental-details-image}, align with full-class in increasing the number of training seeds (rather than unlearning seeds) to reduce \gls{CI} width.

\begin{table}[t]
\centering
\caption{Variance and computational cost of unlearning evaluation in full-class image unlearning. Computational costs are in seconds; unlearning costs include 180~seconds for metric evaluation. Entries show means $\pm$ standard deviations across forget classes and datasets (CIFAR-20 and CIFAR-100). }
\label{tab:variance-components-full-class}
\providecommand{\sd}[1]{{\tiny$\pm#1$}}
\setlength{\tabcolsep}{3pt}
\begin{adjustbox}{max width=\textwidth}
\begin{tabular}{@{} l r@{}l r@{}l r@{}l r@{}l r@{}l r@{}l r@{}l r@{}l @{}}
\toprule
  & &
  & &
  &
  \multicolumn{6}{c}{Forget accuracy}
  & \multicolumn{6}{c}{Retain accuracy} \\
\cmidrule(lr){6-11}\cmidrule(lr){12-17}
{Method}
  & \multicolumn{2}{c}{$\costTrain$}
  & \multicolumn{2}{c}{$\costUnlearn$}
  & \multicolumn{2}{c}{$\varTrainEst$}
  & \multicolumn{2}{c}{$\varUnlearnEst$}
  & \multicolumn{2}{c}{$\iccEst$}
  & \multicolumn{2}{c}{$\varTrainEst$}
  & \multicolumn{2}{c}{$\varUnlearnEst$}
  & \multicolumn{2}{c}{$\iccEst$} \\
\midrule
  SSD (orig.) & $982$ & \sd{840} & $12$ & \sd{2} & $0$ & $.119$\sd{0.375} & $0$ & $.032$\sd{0.100} & $0$ & $.681$\sd{0.153} & $120$ & $.014$\sd{195.181} & $40$ & $.386$\sd{42.348} & $0$ & $.537$\sd{0.438} \\
  SSD (det.) & $1343$ & \sd{828} & $373$ & \sd{38} & $455$ & $.216$\sd{529.057} & $0$ & $.000$\sd{0.000} & $1$ & $.000$\sd{0.000} & $92$ & $.956$\sd{167.487} & $0$ & $.000$\sd{0.000} & $1$ & $.000$\sd{0.000} \\
  LFSSD (orig.) & $982$ & \sd{828} & $12$ & \sd{2} & $0$ & $.000$\sd{0.000} & $0$ & $.000$\sd{0.000} & \multicolumn{2}{c}{---} & $10$ & $.783$\sd{26.213} & $0$ & $.325$\sd{0.703} & $0$ & $.953$\sd{0.012} \\
  LFSSD (det.) & $1344$ & \sd{828} & $374$ & \sd{16} & $0$ & $.000$\sd{0.000} & $0$ & $.000$\sd{0.000} & \multicolumn{2}{c}{---}  & $0$ & $.062$\sd{0.008} & $0$ & $.000$\sd{0.000} & $1$ & $.000$\sd{0.000} \\
  Bad Teacher & $973$ & \sd{829} & $3$ & \sd{1} & $2$ & $.212$\sd{2.926} & $1$ & $.659$\sd{2.401} & $0$ & $.413$\sd{0.235} & $0$ & $.049$\sd{0.004} & $0$ & $.042$\sd{0.008} & $0$ & $.540$\sd{0.042} \\
  UNSIR & $995$ & \sd{840} & $25$ & \sd{12} & $38$ & $.552$\sd{48.156} & $22$ & $.853$\sd{11.398} & $0$ & $.516$\sd{0.242} & $0$ & $.036$\sd{0.004} & $0$ & $.099$\sd{0.046} & $0$ & $.296$\sd{0.098} \\
  Random Labels & $990$ & \sd{829} & $20$ & \sd{1} & $0$ & $.000$\sd{0.000} & $0$ & $.000$\sd{0.001} & $0$ & $.090$\sd{0.000} & $0$ & $.040$\sd{0.008} & $0$ & $.163$\sd{0.096} & $0$ & $.238$\sd{0.100} \\
\bottomrule
\end{tabular}
\end{adjustbox}
\end{table}

\begin{table}[t]
\centering
\caption{Number of training seeds $\nTrainSeeds^*$ and unlearning seeds $\nUnlearnSeeds^*$ per training seed needed to achieve a 95-\% \gls{CI} width $\ciWidth$ in full-class image unlearning, according to Equation~\protect\refeq{eq:ci_sample_size}. Entries show means $\pm$ standard deviations across forget classes and datasets (CIFAR-20 and CIFAR-100).}
\label{tab:optimal-seeds-full-class}
\providecommand{\sd}[1]{{\tiny$\pm#1$}}
\setlength{\tabcolsep}{3pt}
\begin{adjustbox}{max width=\textwidth}
\begin{tabular}{@{} l r@{}l r@{}l r@{}l r@{}l r@{}l r@{}l r@{}l r@{}l r@{}l r@{}l r@{}l r@{}l @{}}
\toprule
  & \multicolumn{12}{c}{Forget accuracy}
  & \multicolumn{12}{c}{Retain accuracy} \\
\cmidrule(lr){2-13}\cmidrule(lr){14-25}
  & \multicolumn{4}{c}{$\ciWidth = 0.5$}
  & \multicolumn{4}{c}{$\ciWidth = 1$}
  & \multicolumn{4}{c}{$\ciWidth = 2$}
  & \multicolumn{4}{c}{$\ciWidth = 0.5$}
  & \multicolumn{4}{c}{$\ciWidth = 1$}
  & \multicolumn{4}{c}{$\ciWidth = 2$} \\
\cmidrule(lr){2-5}\cmidrule(lr){6-9}\cmidrule(lr){10-13}\cmidrule(lr){14-17}\cmidrule(lr){18-21}\cmidrule(lr){22-25}
{Method}
  & \multicolumn{2}{c}{${\nTrainSeeds}^*$}
  & \multicolumn{2}{c}{${\nUnlearnSeeds}^*$}
  & \multicolumn{2}{c}{${\nTrainSeeds}^*$}
  & \multicolumn{2}{c}{${\nUnlearnSeeds}^*$}
  & \multicolumn{2}{c}{${\nTrainSeeds}^*$}
  & \multicolumn{2}{c}{${\nUnlearnSeeds}^*$}
  & \multicolumn{2}{c}{${\nTrainSeeds}^*$}
  & \multicolumn{2}{c}{${\nUnlearnSeeds}^*$}
  & \multicolumn{2}{c}{${\nTrainSeeds}^*$}
  & \multicolumn{2}{c}{${\nUnlearnSeeds}^*$}
  & \multicolumn{2}{c}{${\nTrainSeeds}^*$}
  & \multicolumn{2}{c}{${\nUnlearnSeeds}^*$} \\
\midrule
  SSD (orig.) & $9$ & \sd{26} & $1$ & \sd{0} & $3$ & \sd{6} & $1$ & \sd{0} & $2$ & \sd{2} & $1$ & \sd{0} & $7892$ & \sd{12033} & $4$ & \sd{3} & $1973$ & \sd{3008} & $4$ & \sd{3} & $494$ & \sd{752} & $4$ & \sd{3} \\
  SSD (det.) & $27981$ & \sd{32519} & $1$ & \sd{0} & $6996$ & \sd{8130} & $1$ & \sd{0} & $1749$ & \sd{2032} & $1$ & \sd{0} & $5714$ & \sd{10295} & $1$ & \sd{0} & $1429$ & \sd{2574} & $1$ & \sd{0} & $358$ & \sd{643} & $1$ & \sd{0} \\
  LFSSD (orig.) & $1$ & \sd{0} & $1$ & \sd{0} & $1$ & \sd{0} & $1$ & \sd{0} & $1$ & \sd{0} & $1$ & \sd{0} & $683$ & \sd{1654} & $1$ & \sd{0} & $171$ & \sd{414} & $1$ & \sd{0} & $43$ & \sd{103} & $1$ & \sd{0} \\
  LFSSD (det.) & $1$ & \sd{0} & $1$ & \sd{0} & $1$ & \sd{0} & $1$ & \sd{0} & $1$ & \sd{0} & $1$ & \sd{0} & $4$ & \sd{1} & $1$ & \sd{0} & $1$ & \sd{1} & $1$ & \sd{0} & $1$ & \sd{0} & $1$ & \sd{0} \\
  Bad Teacher & $176$ & \sd{233} & $4$ & \sd{2} & $45$ & \sd{58} & $4$ & \sd{2} & $12$ & \sd{14} & $4$ & \sd{2} & $4$ & \sd{1} & $3$ & \sd{0} & $2$ & \sd{1} & $3$ & \sd{0} & $1$ & \sd{0} & $3$ & \sd{0} \\
  UNSIR & $2941$ & \sd{3402} & $3$ & \sd{1} & $736$ & \sd{851} & $3$ & \sd{1} & $184$ & \sd{213} & $3$ & \sd{1} & $4$ & \sd{1} & $4$ & \sd{1} & $1$ & \sd{0} & $4$ & \sd{1} & $1$ & \sd{0} & $4$ & \sd{1} \\
  Random Labels & $1$ & \sd{0} & $2$ & \sd{2} & $1$ & \sd{0} & $2$ & \sd{2} & $1$ & \sd{0} & $2$ & \sd{2} & $5$ & \sd{1} & $5$ & \sd{1} & $2$ & \sd{1} & $5$ & \sd{1} & $1$ & \sd{0} & $5$ & \sd{1} \\
\bottomrule
\end{tabular}
\end{adjustbox}
\end{table}

\section{Conclusion}

Single-training-seed evaluation of machine unlearning methods, a common practice in the literature, can yield unrepresentative conclusions about their efficacy or robustness; and only using multiple unlearning seeds cannot generally compensate for the lack of multiple training seeds. Our work additionally provides practical guidance on selecting the number of training and unlearning seeds. 

We have illustrated that this issue extends beyond image classification, e.g., to \glspl{LLM}. In this context, we note that using multiple training seeds remains feasible even for \glspl{LLM} (after pre-training). A newly trained model for Llama-3.2-1B-Instruct on TOFU took roughly ten minutes, comparable to ResNet-18.
Future work could include additional tasks such as additional evaluations against more datasets and seeds within the \gls{FOLTR} and \gls{LLM} tasks.

\bibliographystyle{splncs04}
\bibliography{custom}

\appendix
\section{Data and Code Availability}
\label{app:code}

Code for reproducing the experiments -- based on \url{https://github.com/if-loops/selective-synaptic-dampening} (Image), \url{https://github.com/Iris1026/Unlearning-for-FOLTR} (\gls{FOLTR}) with minimal modifications, \url{https://github.com/locuslab/open-unlearning} (\gls{LLM}) with no modifications. Codebase used for running extended Image experiments \url{} \textit{Will be included after Double Blind Review}

\section{Implementation of \gls{SSD} and LFSSD}
\label{sec:det_ssd}

\glsreset{SSD}
\glsreset{LFSSD}

\subsection{Background}

\newcommand{\Df}{D_{\mathrm{f}}}
\newcommand{\Dr}{D_{\mathrm{r}}}

\emph{\Gls{SSD}} \cite{Foster_Schoepf_Brintrup_2024} is a gradient-based unlearning method inspired by \emph{Elastic Weight Consolidation}. The core idea is to dampen (shrink toward zero) a subset of the parameters $\theta = (\theta^1, \dotsc, \theta^d)$ that are important for the forget set $\Df$, but not for the retain set $\Dr$. Importance of $\theta^i$ is measured via the $i$th entry on the diagonal of the empirical Fisher information matrix:
\begin{equation}
    \fisher_\theta^i(D) = \frac{1}{\lvert D \rvert} \sum_{(x,y) \in D} \biggl(\frac{\partial \mathcal{L}(f_\theta(x), y)}{\partial \theta^i}\biggr)^2, \label{eq:empirical_fisher_info}
\end{equation}
where $\mathcal{L}$ denotes a loss function and we assume that data sets consist of feature--label pairs $(x, y)$; $f_\theta$ is the neural network. \gls{SSD} then selects the $i$th parameter for dampening by some factor $\alpha \in [0,1)$ if 
\begin{equation}
    \frac{\fisher_\theta^i(\Df)}{\fisher_\theta^i(\Dr)} \geq \lambda, \label{eq:ssd_ratio}
\end{equation}
where $\lambda \geq 0$ is some threshold chosen by the user.

\emph{\Gls{LFSSD}} \cite{foster2024loss} refines \gls{SSD} by normalising the ratio in Equation~\refeq{eq:ssd_ratio} by dividing it by the maximum of such ratios across the layer to which $\theta_i$ belongs. This prevents layers with naturally large gradient magnitudes from dominating parameter selection, leading to more balanced and targeted forgetting across the network. 

\subsection{Original Implementation -- SSD (orig.) \& LFSSD (orig.)}

The original implementation of \gls{SSD} and \gls{LFSSD} from \cite{Foster_Schoepf_Brintrup_2024,foster2024loss} -- termed \gls{SSD} (orig.) and \gls{LFSSD} (orig.) in this work -- leverages batch computations to perform the per sample parameter importance calculation. 

Specifically, it uses \texttt{reduction="mean"} so that the importance of the $i$th parameter is  computed as:
\begin{equation}
    \tilde{\fisher}_\theta^i(D) = \frac{1}{\lvert \mathcal{B} \rvert} \sum_{B \in \mathcal{B}} \biggl(\frac{1}{\lvert B \rvert} \sum_{(x,y) \in \mathrlap{B}} \frac{\partial \mathcal{L}(f_\theta(x), y)}{\partial \theta^i}\biggr)^2,\label{eq:empirical_fisher_info_biased}
\end{equation}
where $\mathcal{B}$ is the set of batches which make up $D$ (i.e., $\mathcal{B}$ is a partition of $D$). 

In other words, the original implementation (partially) switches the order of averaging and squaring operations. To see this, note that we can express Equation~\refeq{eq:empirical_fisher_info} using an arbitrary batch configuration as
\begin{align}
   \fisher_\theta^i(D).= \frac{1}{\lvert \mathcal{B} \rvert} \sum_{B \in \mathcal{B}} \frac{1}{\lvert B \rvert} \sum_{(x,y) \in B} \biggl(\frac{\partial \mathcal{L}(f_\theta(x), y)}{\partial \theta^i}\biggr)^2.
\end{align}
This has two important consequences:
\begin{enumerate}
    \item The calculated parameter importance used by \gls{SSD} and \gls{LFSSD} is biased since, in general:
\begin{align}
  \tilde{\fisher}_\theta^i(D) \neq 
  \fisher_\theta^i(D).
\end{align}
  \item Both \gls{SSD} and \gls{LFSSD} are not entirely deterministic because the value of $\tilde{\fisher}_\theta^i(D)$ (and hence the above-mentioned bias) depends on the configuration of the batches $\mathcal{B}$; and the latter changes across unlearning seeds (each unlearning seed differently shuffles the batches).
\end{enumerate}

\subsection{Our Implementation -- SSD (det.) \& LFSSD (det.)}

Our experiments in this work (additionally) show results for versions of \gls{SSD} and \gls{LFSSD} which directly use the unbiased importance calculation from Equation~\refeq{eq:empirical_fisher_info}. Note that Equation~\refeq{eq:empirical_fisher_info} does not depend on batch configurations.  We refer to these implementations as \gls{SSD} (det.) and \gls{LFSSD} (det.).

In our experiments, our deterministic implementations improve method stability across training seeds and reduce distance to Retrain.  Results are outlined in Figure~\ref{fig:det-comparison}. 

Follow-up experimentation should be performed to properly tune the parameter $\alpha$ to ensure the deterministic method performs optimally. We have not done this for our experiments, as this is beyond the scope of this work.

\begin{figure}[H]
  \centering

  \resizebox{\linewidth}{!}{\input{tikz/fig_det_comparison.tikz}}

  \caption{Left two charts columns represent full-class results and right two charts represent sub-class results. Top row presents between-seed standard deviation of retain accuracy measures how much outcomes vary across different unlearning seeds (i.e., different random shuffles of the importance-computation dataloaders). Middle row presents within-seed standard deviation measures run-to-run variance under a fixed initial seed. Implementation reduces this to zero demonstrating the methods are now deterministic. Bottom row presents mean absolute distance from the retrain baseline. Error bars show the standard deviation across training seeds and classes; pp in the second-axis label denotes percentage points.}
  \label{fig:det-comparison}
\end{figure}

\section{Additional Experimental Results -- Image}
\label{app:experimental-details-image}

This section provides comprehensive details on the experimental configurations and hyperparameters for image classification tasks as well as sub-class results. All experiments are performed using a single NVIDIA L40S GPU (48GB) on a SLURM based HPC cluster.


\subsection{Model Architecture and Training}

We use a ResNet-18 architecture, randomly initialised and trained from scratch using stochastic gradient descent (SGD) with momentum $0.9$ and weight decay $5 \times 10^{-4}$. The initial learning rate is $0.1$, decayed by a factor of $0.2$ at predefined milestones via a multi-step learning rate scheduler. A single epoch of linear warmup is applied at the start of training. The batch size is $64$ for all experiments. Input images are $32 \times 32$ pixels for CIFAR datasets. Dataset-specific training schedules are listed in Table~\ref{tab:vision-training-schedule}. We trained 75 initial models ($\nTrainSeeds = 75$), initial seeds and ran unlearning experiments for each across 10 unlearning seeds ($\nUnlearnSeeds = 10$).

\begin{table}[h]
\centering
\caption{Training schedules for image classification datasets (ResNet-18).}
\label{tab:vision-training-schedule}
{\begin{tabular}{lrrr}
\toprule
Dataset & Classes & Epochs & LR Milestones \\
\midrule
CIFAR-10  & 10  & 20  & [8, 12, 16] \\
CIFAR-20  & 20  & 40  & [15, 30, 35] \\
CIFAR-100 & 100 & 200 & [60, 120, 160] \\
\bottomrule
\end{tabular}}
\end{table}

The following RNG-level sources of randomness (Python \emph{random}, NumPy, PyTorch, and dataloader workers) are seeded. Models are trained at \emph{fp32} (\emph{32-true}) precision. Kernel-level seeding and determinism was not enabled \\(\emph{torch.use\_deterministic\_algorithms}, \emph{torch.backends.cudnn.deterministic}, \\ \emph{CUBLAS\_WORKSPACE\_CONFIG}); \emph{torch.backends.cudnn.benchmark} was left at its default of \emph{False}, which fixes the choice of convolution algorithm but not the determinism of the chosen kernel. Consequently, the following operations in our pipeline use non-deterministic CUDA kernels in their backward pass: \emph{nn.Conv2d} (the ResNet-18 backbone), \emph{nn.AdaptiveAvgPool2d} (the global pooling layer, which has no deterministic CUDA implementation),\\ \emph{nn.CrossEntropyLoss} (via \emph{NLLLoss}), and \emph{nn.Linear} (via cuBLAS matrix multiplication, non-deterministic for CUDA $\geq 10.2$ unless \emph{CUBLAS\_WORKSPACE} \emph{\_CONFIG} is set). Forward passes and the resulting accuracy computation (a forward pass followed by an \emph{argmax}) is deterministic, so non-deterministic behaviour is confined to gradient computations.

\subsection{Evaluation Metrics}

We evaluate unlearning quality using the following metrics:
\begin{itemize}
    \item \textbf{Retain accuracy}: Classification accuracy on the test samples of the retained classes.
    \item \textbf{Forget accuracy}: Classification accuracy on the test samples of the forgotten class
\end{itemize}

\subsection{Unlearning Method Hyperparameters}

\begin{itemize}
    \item SSD (orig.): $\lambda = 1$, $\alpha = 10$, Dampening constant $\lambda$ and selection weighting $\alpha$
    \item LFSSD (orig.): $\lambda = 1$, $\alpha = 10$, Same as \gls{SSD} but uses loss-free importance calculation; $\alpha$ = $5$ for CIFAR-20 
    \item \gls{SSD} (det.): Same as \gls{SSD} but with adjusted implementation \ref{sec:det_ssd}.
    \item \gls{LFSSD} (det.):  Same as \gls{LFSSD} but with adjusted implementation \ref{sec:det_ssd}. 
    \item Bad Teacher: LR $= 0.0001$, epochs $= 1$, KL temperature $= 1$, Adam optimiser; batch size $= 256$
    \item UNSIR: LR $= 0.0001$, noise epochs $= 25$, batch size $= 32$, Adam optimiser; 500 retain samples per class
    \item Random Labels: LR $= 0.0001$, epochs $= 3$, batch size $= 128$,  Adam optimiser; forget-class samples relabelled uniformly at random from remaining classes; trained jointly with retain set 
\end{itemize}

\subsection{Sub-Class Results}

Similar to results from full-class unlearning, many unlearning methods exhibit high training-seed variation, as shown in Table~\ref{tab:variance-components-sub-class}. However, there are several methods, such as Bad Teacher and UNSIR, whose forget accuracy $\varUnlearn$ was comparable to $\varTrain$ and in the case of Random Labels was substantially higher. As shown in Table~\ref{tab:optimal-seeds-sub-class} this highlights the need for not fully discounting unlearning seeds as part of MU method evaluation however the results still consistently show that training seeds are more likely to be needed to capture the variance of MU methods.

\begin{table}[t]
\centering
\caption{Variance and computational cost of unlearning evaluation in sub-class image unlearning. Computational costs are in seconds; unlearning costs include 180~seconds for metric evaluation. Entries show means $\pm$ standard deviations across forget classes.}
\label{tab:variance-components-sub-class}
\providecommand{\sd}[1]{{\tiny$\pm#1$}}
\setlength{\tabcolsep}{3pt}
\begin{adjustbox}{max width=\textwidth}
\begin{tabular}{@{} l r@{}l r@{}l r@{}l r@{}l r@{}l r@{}l r@{}l r@{}l @{}}
\toprule
  & &
  & &
  &
  \multicolumn{6}{c}{Forget accuracy}
  & \multicolumn{6}{c}{Retain accuracy} \\
\cmidrule(lr){6-11}\cmidrule(lr){12-17}
{Method}
  & \multicolumn{2}{c}{$\costTrain$}
  & \multicolumn{2}{c}{$\costUnlearn$}
  & \multicolumn{2}{c}{$\varTrainEst$}
  & \multicolumn{2}{c}{$\varUnlearnEst$}
  & \multicolumn{2}{c}{$\iccEst$}
  & \multicolumn{2}{c}{$\varTrainEst$}
  & \multicolumn{2}{c}{$\varUnlearnEst$}
  & \multicolumn{2}{c}{$\iccEst$} \\
\midrule
  SSD (orig.) & $982$ & \sd{840} & $12$ & \sd{2} & $6$ & $.157$\sd{13.760} & $0$ & $.715$\sd{1.594} & $0$ & $.729$\sd{0.237} & $22$ & $.678$\sd{44.592} & $1$ & $.173$\sd{2.234} & $0$ & $.927$\sd{0.033} \\
  SSD (det.) & $1343$ & \sd{828} & $373$ & \sd{38} & $295$ & $.127$\sd{198.874} & $0$ & $.000$\sd{0.000} & $1$ & $.000$\sd{0.000} & $0$ & $.608$\sd{0.497} & $0$ & $.000$\sd{0.000} & $1$ & $.000$\sd{0.000} \\
  LFSSD (orig.) & $982$ & \sd{828} & $12$ & \sd{2} & $0$ & $.071$\sd{0.158} & $0$ & $.008$\sd{0.018} & $0$ & $.899$\sd{0.000}  & $10$ & $.695$\sd{18.979} & $0$ & $.248$\sd{0.418} & $0$ & $.960$\sd{0.022} \\
  LFSSD (det.) & $1344$ & \sd{828} & $374$ & \sd{16} & $43$ & $.563$\sd{36.943} & $0$ & $.000$\sd{0.000} & $1$ & $.000$\sd{0.000} & $0$ & $.120$\sd{0.040} & $0$ & $.000$\sd{0.000} & $1$ & $.000$\sd{0.000} \\
  Bad Teacher & $973$ & \sd{829} & $3$ & \sd{1} & $7$ & $.608$\sd{4.248} & $11$ & $.971$\sd{6.367} & $0$ & $.385$\sd{0.023} & $0$ & $.045$\sd{0.004} & $0$ & $.037$\sd{0.002} & $0$ & $.547$\sd{0.029} \\
  UNSIR & $995$ & \sd{840} & $25$ & \sd{12} & $32$ & $.203$\sd{26.966} & $27$ & $.712$\sd{19.817} & $0$ & $.484$\sd{0.170} & $0$ & $.793$\sd{0.312} & $0$ & $.510$\sd{0.061} & $0$ & $.592$\sd{0.090} \\
  Random Labels & $990$ & \sd{829} & $20$ & \sd{1} & $7$ & $.176$\sd{8.835} & $43$ & $.404$\sd{56.107} & $0$ & $.182$\sd{0.046} & $0$ & $.035$\sd{0.003} & $0$ & $.065$\sd{0.003} & $0$ & $.346$\sd{0.022} \\
\bottomrule
\end{tabular}
\end{adjustbox}
\end{table}

\begin{table}[t]
\centering
\caption{Number of training seeds $\nTrainSeeds^*$ and unlearning seeds $\nUnlearnSeeds^*$ per training seed needed to achieve a 95-\%-\gls{CI} width $\ciWidth$ in sub-class image unlearning, according to Equation~\protect\refeq{eq:ci_sample_size}. Entries show means $\pm$ standard deviations across forget classes.}
\label{tab:optimal-seeds-sub-class}
\providecommand{\sd}[1]{{\tiny$\pm#1$}}
\setlength{\tabcolsep}{3pt}
\begin{adjustbox}{max width=\textwidth}
\begin{tabular}{@{} l r@{}l r@{}l r@{}l r@{}l r@{}l r@{}l r@{}l r@{}l r@{}l r@{}l r@{}l r@{}l @{}}
\toprule
  & \multicolumn{12}{c}{Forget accuracy}
  & \multicolumn{12}{c}{Retain accuracy} \\
\cmidrule(lr){2-13}\cmidrule(lr){14-25}
  & \multicolumn{4}{c}{$\ciWidth = 0.5$}
  & \multicolumn{4}{c}{$\ciWidth = 1$}
  & \multicolumn{4}{c}{$\ciWidth = 2$}
  & \multicolumn{4}{c}{$\ciWidth = 0.5$}
  & \multicolumn{4}{c}{$\ciWidth = 1$}
  & \multicolumn{4}{c}{$\ciWidth = 2$} \\
\cmidrule(lr){2-5}\cmidrule(lr){6-9}\cmidrule(lr){10-13}\cmidrule(lr){14-17}\cmidrule(lr){18-21}\cmidrule(lr){22-25}
{Method}
  & \multicolumn{2}{c}{${\nTrainSeeds}^*$}
  & \multicolumn{2}{c}{${\nUnlearnSeeds}^*$}
  & \multicolumn{2}{c}{${\nTrainSeeds}^*$}
  & \multicolumn{2}{c}{${\nUnlearnSeeds}^*$}
  & \multicolumn{2}{c}{${\nTrainSeeds}^*$}
  & \multicolumn{2}{c}{${\nUnlearnSeeds}^*$}
  & \multicolumn{2}{c}{${\nTrainSeeds}^*$}
  & \multicolumn{2}{c}{${\nUnlearnSeeds}^*$}
  & \multicolumn{2}{c}{${\nTrainSeeds}^*$}
  & \multicolumn{2}{c}{${\nUnlearnSeeds}^*$}
  & \multicolumn{2}{c}{${\nTrainSeeds}^*$}
  & \multicolumn{2}{c}{${\nUnlearnSeeds}^*$} \\
\midrule
  SSD (orig.) & $423$ & \sd{944} & $1$ & \sd{1} & $106$ & \sd{236} & $1$ & \sd{1} & $27$ & \sd{59} & $1$ & \sd{1} & $1466$ & \sd{2878} & $1$ & \sd{0} & $367$ & \sd{719} & $1$ & \sd{0} & $92$ & \sd{180} & $1$ & \sd{0} \\
  SSD (det.) & $18141$ & \sd{12224} & $1$ & \sd{0} & $4535$ & \sd{3056} & $1$ & \sd{0} & $1134$ & \sd{764} & $1$ & \sd{0} & $38$ & \sd{30} & $1$ & \sd{0} & $10$ & \sd{8} & $1$ & \sd{0} & $3$ & \sd{2} & $1$ & \sd{0} \\
  LFSSD (orig.) & $6$ & \sd{11} & $1$ & \sd{0} & $2$ & \sd{3} & $1$ & \sd{0} & $1$ & \sd{0} & $1$ & \sd{0} & $673$ & \sd{1192} & $1$ & \sd{0} & $169$ & \sd{298} & $1$ & \sd{0} & $43$ & \sd{75} & $1$ & \sd{0} \\
  LFSSD (det.) & $2678$ & \sd{2271} & $1$ & \sd{0} & $670$ & \sd{568} & $1$ & \sd{0} & $168$ & \sd{142} & $1$ & \sd{0} & $8$ & \sd{2} & $1$ & \sd{0} & $2$ & \sd{1} & $1$ & \sd{0} & $1$ & \sd{0} & $1$ & \sd{0} \\
  Bad Teacher & $679$ & \sd{377} & $4$ & \sd{1} & $170$ & \sd{94} & $4$ & \sd{1} & $43$ & \sd{23} & $4$ & \sd{1} & $4$ & \sd{0} & $3$ & \sd{0} & $1$ & \sd{0} & $3$ & \sd{0} & $1$ & \sd{0} & $3$ & \sd{0} \\
  UNSIR & $2551$ & \sd{2044} & $3$ & \sd{1} & $638$ & \sd{511} & $3$ & \sd{1} & $160$ & \sd{128} & $3$ & \sd{1} & $64$ & \sd{22} & $2$ & \sd{0} & $16$ & \sd{6} & $2$ & \sd{0} & $5$ & \sd{1} & $2$ & \sd{0} \\
  Random Labels & $893$ & \sd{1112} & $5$ & \sd{1} & $224$ & \sd{278} & $5$ & \sd{1} & $56$ & \sd{69} & $5$ & \sd{1} & $4$ & \sd{1} & $4$ & \sd{0} & $1$ & \sd{0} & $4$ & \sd{0} & $1$ & \sd{0} & $4$ & \sd{0} \\
\bottomrule
\end{tabular}
\end{adjustbox}
\end{table}

\subsection{Extended Version of Figure~\ref{fig:intro_figure_trimmed}}
\label{app:subsec:extended_version_of_intro_figure}

Figure~\ref{fig:intro_figure} gives an extended version of Figure~\ref{fig:intro_figure_trimmed} from the main paper -- without any trimming of the second axis and with further separation of the results into those for full-class and sub-class unlearning. Specifically, in this figure, each boxplot visualises either $10$ estimates (for full-class unlearning on CIFAR-100 and CIFAR-20 each with five forget classes -- Panel~\ref{fig:intro_figure:full_class}) or $5$ estimates (for sub-class unlearning on CIFAR-100 with five forget classes -- Panel~\ref{fig:intro_figure:sub_class}) of the standard deviation of the retain and forget accuracies (in percentage points): overall ($\sdOverallEst$), across training seeds ($\sdTrainEst$), and across unlearning seeds ($\sdUnlearnEst$). Each of these $10$ or $5$ estimates is based on 75 training seeds; each training seed is combined with ten unlearning seeds.

\begin{figure}[!htb]
    \centering
    \begin{subfigure}{\linewidth}
    \input{tikz/intro_figure_full_class.tex}
        \vspace{-2ex}
    \caption{Full-class unlearning (CIFAR-100 and CIFAR-20).}
    \label{fig:intro_figure:full_class}
  \end{subfigure}
    \begin{subfigure}{\linewidth}
    \input{tikz/intro_figure_sub_class.tex}
        \vspace{-2ex}
    \caption{Sub-class unlearning (CIFAR-100).}
    \label{fig:intro_figure:sub_class}
  \end{subfigure}
    \caption{Variability unlearning performance in the image classification scenario from Section~\ref{sec:experiments} (extended version of Figure~\ref{fig:intro_figure_trimmed}).
    }
    \label{fig:intro_figure}
\end{figure}

\section{Additional Experimental Results -- FOLTR}
\label{app:foltr-details}

This section provides comprehensive details on the experimental configurations and hyperparameters for federated online learning-to-rank (FOLTR) tasks. All experiments are performed on a single NVIDIA A100-SXM4-80GB GPU.

\subsection{Dataset}

Following \cite{foltrrepTao} we use the MQ2007 dataset from the LETOR 4.0 benchmark. MQ2007 contains 1,692 queries with 46 features per query-document pair and relevance judgements on a three-point scale ($\{0, 1, 2\}$). We use all 5 standard folds, training on the training split and evaluating on the corresponding test split for each fold. No query-level normalisation is applied for MQ2007.

\subsection{Model Architecture}

We use a linear ranker with 46 input features (matching MQ2007's feature dimensionality), randomly initialised. The ranker uses a softmax temperature $\tau = 1$ and learning rate decay factor of $1$ (no decay). Ranking is performed over the top-$k$ documents where $k = \min\{10, n_{\text{docs}}\}$.

\subsection{Federated Training Configuration}

The federated training configuration. Outlined in Table \ref{tab:foltr-hyperparams}

\begin{table}[h]
\centering
\caption{FOLTR training hyperparameters.}
\label{tab:foltr-hyperparams}
{\begin{tabular}{lc}
\toprule
\textbf{Parameter} & \textbf{Value} \\
\midrule
Number of clients & 10 \\
Learning rate & 0.1 \\
Interactions per feedback round & 5 \\
Total interaction budget & 50{,}000 \\
Training iterations & 1{,}000 \\
Aggregation method & Federated Averaging \\
Client data distribution & Query-level partitioning \\
Number of malicious clients & 3 (for poisoning scenarios) \\
Online discount factor & 0.9995 \\
\bottomrule
\end{tabular}}
\end{table}

Each federated round proceeds as follows: (1) each client performs local ranking updates using $5$ user interactions with the update rule; (2) clients send their updated weights to the server; (3) the server aggregates via federated averaging; (4) the updated global model is broadcast back to all clients. Offline NDCG@10 is evaluated on the held-out test set after every round.

The total number of training iterations is $50{,}000 / (10 \times 5) = 1{,}000$. Unlearning then proceeds for an additional $1{,}000$ iterations (matching the training budget), giving $2{,}000$ total epochs as shown in the figures.

\subsection{Click Models}
\label{app:click-models}

User interactions are simulated using a Cascade Click Model with examination depth 10. Three user behaviour profiles are used, parameterised by click relevance $P(\text{click} \mid \text{rel})$ and stop relevance $P(\text{stop} \mid \text{rel})$ as shown in Table~\ref{tab:click-models}.

\begin{table}[h]
\centering
\caption{Click model parameters for MQ2007 (3-point relevance scale: $\{0, 1, 2\}$).}
\label{tab:click-models}
{\begin{tabular}{lcccccc}
\toprule
 & \multicolumn{3}{c}{$P(\text{click} \mid \text{rel})$} & \multicolumn{3}{c}{$P(\text{stop} \mid \text{rel})$} \\
\cmidrule(lr){2-4} \cmidrule(lr){5-7}
\textbf{Model} & $r{=}0$ & $r{=}1$ & $r{=}2$ & $r{=}0$ & $r{=}1$ & $r{=}2$ \\
\midrule
Perfect        & 0.0  & 0.5  & 1.0  & 0.0 & 0.0 & 0.0 \\
Navigational   & 0.05 & 0.5  & 0.95 & 0.2 & 0.5 & 0.9 \\
Informational  & 0.4  & 0.7  & 0.9  & 0.1 & 0.3 & 0.5 \\
\bottomrule
\end{tabular}}
\end{table}

The \emph{Perfect} model provides noise-free relevance feedback. The \emph{Navigational} model represents users seeking a single highly relevant document (high stop probability at $r{=}2$). The \emph{Informational} model represents users examining multiple documents with noisier click patterns. For the \textbf{data poisoning} scenario, malicious clients use a \emph{Poison} click model with inverted relevance: $P(\text{click} \mid r{=}0) = 1.0$, $P(\text{click} \mid r{=}1) = 0.5$, $P(\text{click} \mid r{=}2) = 0.0$, and all stop probabilities set to $0.0$.

\subsection{Attack Scenarios}

We evaluate under three scenarios:
\begin{itemize}
    \item \textbf{Clean}: All 10 clients use the same honest click model. No attack is present.
    \item \textbf{Data poisoning}: 3 of 10 clients use the Poison click model (inverted relevance), while the remaining 7 use the honest click model. The server is unaware of which clients are malicious.
    \item \textbf{Model poisoning}: 3 of 10 clients are flagged as malicious and send adversarially crafted weight updates (e.g., shuffled document scores), while using the honest click model for local interactions.
\end{itemize}

\subsection{Results}
In \gls{FOLTR} multiple clients collectively train a ranking model to rank documents for search queries while learning from simulated user interactions (clicks). Unlearning aims to remove the influence where malicious clients or incorrect data that poisons the federated ranking model. The following unlearning methods were examined: \emph{finetune} \cite{finetuneFOLTRZhang2023}, \emph{Gradient Ascent} \cite{halimi2023federatedunlearningefficientlyeraseGA}, \emph{FedRemove} \cite{FedRemoveYuan2023} and \emph{FedEraser} \cite{FedEraser2021Liu}. 

We compare the common practice of using only a single training seed, specifically: $(\nTrainSeeds, \nUnlearnSeeds) = (1, 10)$,  with $(\nTrainSeeds, \nUnlearnSeeds) = (10, 1)$ (Figure~\ref{fig:fediroltrunlearning}). We use the MQ2007 dataset. A linear ranker was randomly initialised and trained from scratch in a federated online learning setting with 10 clients on the dataset. 

Three click behaviour models (Perfect, Navigational, and Informational) were employed to simulate different user interaction patterns, following \cite{foltrrepTao}. Evaluation is performed as demonstrated in Figure~\ref{fig:fediroltrunlearning} using normalised discounted cumulative gain at position 10 (NDCG@10), which measures ranking quality by comparing the system's ranking to the ideal ranking based on document relevance labels. NDCG@10 is normalised to $[0, 1]$, where higher values indicate better unlearning performance.

\begin{figure}[H]
\centering

\includegraphics[width=\linewidth]{images/MQ2007_model_poison_algorithm2.png}

\caption{FOLTR on MQ2007 model poison unlearning scenario across $I=10$ training seeds ($J=1$ unlearning seed each). Offline NDCG@10 (evaluated on held-out test set after each training/unlearning epoch), scores are plotted over 2000 epochs. The first 1000 epochs represent the training phase, epochs 1000-2000 represent the unlearning phase. Thin transparent lines show individual, runs thick solid lines represent the mean trajectory and shaded regions represent standard deviation. }
\label{fig:fediroltrunlearning}
\end{figure}

The results presented in Figure~\ref{fig:fediroltrunlearning} corroborate the findings from the image classification experiments. Evaluating unlearning with a single training seed (i.e., $I=1$, combined with multiple unlearning seeds) yields trajectories whose spread is markedly non-representative of the true cross-seed variability. 

In contrast, evaluating across multiple training seeds (each paired with a single unlearning seed) exposes substantially greater variance in post-unlearning performance. The seed sensitivity observed for unlearning methods in image classification extends to the ranking domain. This outcome is consistent across all three simulated user-interaction patterns, indicating that the limitation of single-training-seed evaluation is not an artefact of a particular click model, but a general property of the evaluation framework.

\section{Additional Experimental Results -- LLM}
\label{app:LLM-details}

\subsection{Experiment Outline}
Utilising the open unlearning framework \cite{openunlearning2025} we ran a small set of experiments on the TOFU \cite{maini2024tofu} dataset unlearning scenario using Llama-3.2-1B-Instruct as the architecture loaded in bfloat16 precision with flash attention. Each initial model is trained in an identical way to the model provided by the framework authors, henceforth refereed to by as the \emph{remote} model. All methods evaluated are those included within the open unlearning framework.

\subsection{Evaluation Metrics}

We evaluate unlearning quality using two aggregate metrics from the TOFU benchmark:

\begin{itemize}
    \item \textbf{Model Utility}: The harmonic mean of retain-set performance metrics, measuring how well the model preserves general language modelling capability after unlearning. Higher is better.
    \item \textbf{Forget Quality}: The $p$-value from a Kolmogorov--Smirnov (KS) test comparing the distribution of forget-set outputs produced by the unlearned model against those of the retain model. A higher $p$-value indicates that the unlearned model's behaviour on the forget set is statistically indistinguishable from the retain model, signifying more complete forgetting.
\end{itemize}

\subsection{Results}
\begin{figure}[H]
\centering
\includegraphics[width=\linewidth]{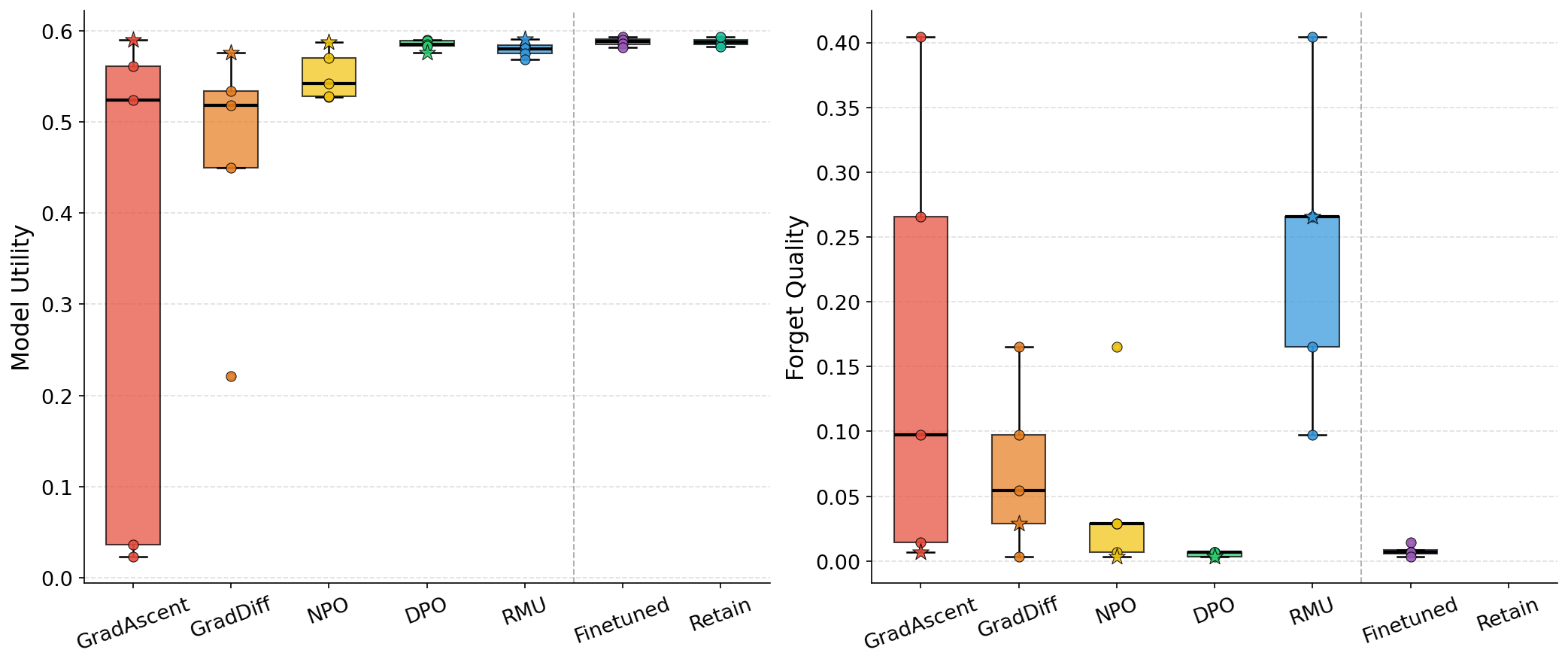}
\caption{TOFU forget01 unlearning results for Llama-3.2-1B-Instruct across $I=5$ training seeds ($J=1$ unlearning seed each). Left: Model Utility (harmonic mean of retain-set metrics; higher is better). Right: Forget Quality (KS $p$-value vs.\ retain model; higher indicates better forgetting). Individual seeds are shown as points; star markers denote the remote seed. Finetune and Retain are shown as baselines (right of dashed line, $I=4$ seeds). Starred points are data points gathered by unlearning using remote seed provided, all our data points plotted as dots. }
\label{fig:llm-unlearning}
\end{figure}

The results in Figure~\ref{fig:llm-unlearning} extend our findings from the vision and \gls{FOLTR} domains to the \gls{LLM} setting. Substantial cross-seed variance is visible even within a some methods such as GradAscent and Runlearning, these exhibit wide distributions across both Model Utility and Forget Quality, with individual seeds spanning nearly the full range of possible outcomes. This initial seed sensitivity demonstrates that evaluating with a single training seed can be insufficient to characterise the behaviour of an unlearning method in the unlearning \gls{LLM} setting, consistent with our findings across all three experimental domains.

\section{Details on the Variance Decomposition}
\label{app:sec:variance_decomposition}

In this section, we derive the variance decomposition from Equation~\refeq{eq:variance_decomposition}.

Let $\TrainSeed_i$ and $\UnlearnSeed_{i,j}$ be randomly and independently, chosen training and unlearning seeds, respectively. Further, for some given architecture, unlearning method, and retain and forget sets, let $\metric_{i,j} = h(\TrainSeed_i, \UnlearnSeed_{i,j})$, where $h(\trainSeed, \unlearnSeed) \coloneqq \evaluate(\unlearn(\train(\seed = \trainSeed), \seed = \unlearnSeed))$, be the value of some metric for assessing unlearning under a model first trained using training seed $\TrainSeed_i$ and then unlearned using unlearning seed $\UnlearnSeed_{i,j}$ (e.g., $\metric_{i,j}$ could be the accuracy on the forget set or retain set).   

Then by independence of the training and unlearning seeds, using the standard variance decomposition:
\begin{align}
  \var[\metricAvg]
  & = \var\biggl[ \frac{1}{\nTrainSeeds \nUnlearnSeeds} \sum_{i=1}^{\nTrainSeeds} \sum_{j=1}^{\nUnlearnSeeds} Z_{i,j}\biggr]\\
  & = \var\biggl[ \frac{1}{\nTrainSeeds \nUnlearnSeeds}  \sum_{i=1}^{\nTrainSeeds} \sum_{j=1}^{\nUnlearnSeeds} h(\TrainSeed_i, \UnlearnSeed_{i,j})\biggr]\\
  & = \frac{1}{\nTrainSeeds \nUnlearnSeeds^2}   \var\biggl[ \sum_{j=1}^{\nUnlearnSeeds} h(\TrainSeed_1, \UnlearnSeed_{1,j})\biggr]\\
 & = \frac{1}{\nTrainSeeds \nUnlearnSeeds^2}   \biggl(\var\biggl[\E\biggl(\sum_{j=1}^{\nUnlearnSeeds} h(\TrainSeed_1, \UnlearnSeed_{1,j})\,\bigg|\, \TrainSeed_1\biggr) \biggr]\\
 & \qquad + \E\biggl[\var\biggl(\sum_{j=1}^{\nUnlearnSeeds} h(\TrainSeed_1, \UnlearnSeed_{1,j})\,\bigg|\, \TrainSeed_1\biggr) \biggr]\biggr)\\
  & = \frac{1}{\nTrainSeeds}  \var[\E(h(\TrainSeed_1, \UnlearnSeed_{1,1}) | \TrainSeed_1) ] + \frac{1}{\nTrainSeeds \nUnlearnSeeds} \E[\var(h(\TrainSeed_1, \UnlearnSeed_{1,1}) | \TrainSeed_1)].
\end{align}
\end{document}